\documentclass{article}

\usepackage{arxiv}

\usepackage[utf8]{inputenc} 
\usepackage[T1]{fontenc}    
\usepackage{nicefrac}       
\usepackage{microtype}      
\usepackage{amsmath}
\usepackage{amssymb}
\usepackage{amsfonts}
\usepackage{amsthm}         

\usepackage{graphicx}
\usepackage{booktabs}       
\usepackage{multirow}
\usepackage[table]{xcolor}
\usepackage[caption=false,font=normalsize,labelfont=sf,textfont=sf]{subfig}
\usepackage{makecell}

\usepackage[ruled,vlined,linesnumbered]{algorithm2e}
\SetKwInput{KwInput}{Input}
\SetKwInput{KwOutput}{Output}
\SetKw{KwInit}{Initialize}
\SetKw{KwRet}{return}
\SetKw{KwAnd}{and}
\SetKw{KwOr}{or}

\usepackage{url}          
\usepackage{natbib}
\let\cite\citep
\usepackage{doi}
\usepackage{hyperref}  

\usepackage{pgfplots}
\usepackage{pgfplotstable}
\usepgfplotslibrary{groupplots}
\pgfplotsset{compat=1.18}
\usepackage{xcolor}
\usepackage{tikz}

\usepackage[shortcuts,acronym]{glossaries}
\glsdisablehyper
\newif\ifuniqueAffiliation
\uniqueAffiliationfalse

\usepackage{authblk}

\newbox{\orcid}\sbox{\orcid}{\includegraphics[scale=0.06]{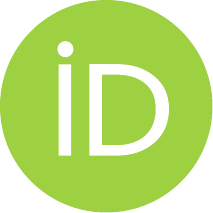}} 

\title{Latent Diffusion for Internet of Things Attack Data Generation in Intrusion Detection \thanks{This work has been submitted to the IEEE for possible publication}}

\date{}

\author[1]{%
    \href{https://orcid.org/0009-0000-5401-9805}{\usebox{\orcid}\hspace{1mm}Estela~Sánchez-Carballo\thanks{\texttt{estela.sanchezc@urjc.es}}}%
}
\author[1]{%
    \href{https://orcid.org/0000-0001-6916-6082}{\usebox{\orcid}\hspace{1mm}Francisco~M.~Melgarejo-Meseguer}%
}
\author[1,2]{%
    \href{https://orcid.org/0000-0003-0426-8912}{\usebox{\orcid}\hspace{1mm}José~Luis~Rojo-Álvarez}%
}

\affil[1]{Departamento de Teoría de la Señal y Comunicaciones y Sistemas Telemáticos y Computación, \protect\\ Universidad Rey Juan Carlos, Madrid, Spain.}
\affil[2]{Data, Networks and Cybersecurity Research Institute, Universidad Rey Juan Carlos, Madrid, Spain.}
\renewcommand{\shorttitle}{Latent Diffusion for Internet of Things Attack Data Generation in Intrusion Detection}

\hypersetup{
pdftitle={Latent Diffusion for Internet of Things Attack Data Generation in Intrusion Detection},
pdfsubject={cs.LG},
pdfauthor={E.~Sánchez-Carballo et al.},
pdfkeywords={Data Augmentation, Diffusion Models, Internet of Things, Intrusion Detection Systems, Latent Spaces, Tabular Data Synthesis.},
}

\begin{document}
\maketitle

\begin{abstract}
The growing deployment of Internet of Things systems has increased the need for reliable intrusion detection under heterogeneous and imbalanced network traffic. However, representative malicious traffic, particularly for underrepresented attacks, remains difficult to obtain, while existing generative approaches do not consistently preserve distributional fidelity, multivariate dependencies, and coverage simultaneously. This work proposes Latent Diffusion for Intrusion Synthesis (LaDIS), an attack-specific framework that learns a mixed-type autoencoder representation and performs diffusion in an adaptive continuous latent space. LaDIS was compared with established adversarial, variational, and direct-space diffusion approaches across representative Flooding, Botnet, Recon, and Stealthy attacks using complementary measures of marginal and multivariate fidelity, dependency preservation, real--synthetic distinguishability, manifold structure, and computational efficiency. LaDIS provided a competitive balance across these dimensions, with particularly strong results for representative Botnet, Recon, and Stealthy attacks, while revealing an attack-dependent fidelity--coverage trade-off. Performing diffusion in latent space also reduced sampling time by approximately 3--5 times relative to direct-space diffusion, with negligible decoding overhead. Downstream utility was evaluated using TabNet and XGBoost, demonstrating that partial LaDIS augmentation consistently produced the best performance, increasing balanced accuracy from 0.86 to 0.94 for TabNet and from 0.92 to 0.96 for XGBoost, while improving the F1-score for low-volume attack families from 0.73 to 0.84 and from 0.82 to 0.88, respectively. These results show that attack-adaptive latent diffusion can generate structurally faithful attack traffic efficiently and, when used selectively, improve intrusion detection for underrepresented attacks.
\end{abstract}



\section{Introduction}

The rapid proliferation of Internet of Things (IoT) devices has transformed domains such as healthcare, smart cities, industrial automation, and intelligent transportation \cite{Perwej19, Lohiya21}, while simultaneously expanding the attack surface of modern networks \cite{Shafiq22}. Consequently, intrusion detection systems (IDSs) have become essential for protecting IoT infrastructures \cite{Khraisat19}. Machine learning (ML)- and deep learning (DL)-based IDSs have shown strong performance by learning complex patterns from network traffic \cite{Meseguer21,Huang25}. However, IoT intrusion detection datasets are often highly imbalanced, with benign traffic and certain high-volume attacks accounting for a large proportion of observations, while other attack types remain underrepresented \cite{Ma25, Mallidi25}. This imbalance can hinder the detection of minority classes and limit model generalization. In this setting, data augmentation offers a promising strategy to address this limitation by increasing the representation of scarce attack patterns during IDS training.

Several data augmentation strategies can be used to mitigate data scarcity and class imbalance. Conventional approaches such as random oversampling are simple and computationally inexpensive, but repeatedly presenting the same minority samples may promote overfitting and does not increase the diversity of the training distribution \cite{Azhar22}. More recently, deep generative models, including generative adversarial networks (GANs), variational autoencoders (VAEs), and diffusion-based approaches, have emerged as alternatives for synthesizing new observations, with several architectures specifically designed for tabular data \cite{LeiXu19,Kotelnikov23}. Each paradigm presents different strengths and limitations. GANs may suffer from training instability and limited mode coverage, VAEs can produce overly smooth distributions, and diffusion models (DMs) can be computationally demanding \cite{Yazdani25}. These limitations are particularly relevant in network traffic generation, where realistic synthetic samples must reproduce not only individual feature distributions, but also the multivariate structure, inter-feature dependencies, and diversity of traffic and attack patterns observed in real data \cite{XuanruiXiong26}.

A promising strategy to reduce the complexity of generative modeling is to first learn a compact representation of the data and perform diffusion in this latent space \cite{Zhang24,Zhu25}. Operating on a lower-dimensional representation can improve computational efficiency while allowing the generative process to focus on the underlying structure of the traffic. However, the effectiveness of this approach depends on how well the latent representation, typically learned using an autoencoder (AE), captures the underlying structure of the data. Excessive compression may discard relevant information, whereas unnecessarily large latent spaces may reduce the advantages of latent modeling \cite{Saha25}. Motivated by this trade-off, we propose Latent Diffusion for Intrusion Synthesis (LaDIS), a latent diffusion framework that learns an attack-specific representation, performs the diffusion process in this latent domain, and reconstructs the generated samples back into the original feature space. LaDIS incorporates a configurable latent dimensionality, adapting to each attack while aiming to preserve both marginal feature distributions and multivariate dependencies.

The main contributions of this work are: (i) LaDIS, a latent diffusion framework for synthetic IoT attack generation that combines mixed-type AE representations with diffusion in a compact continuous latent space; (ii) an attack-specific generative modeling strategy in which independent models are learned for individual attack types rather than relying on a single class-conditioned generator, enabling targeted synthesis while preserving attack-specific traffic characteristics; (iii) an attack-adaptive latent dimensionality selection strategy that configures representation capacity independently for each attack according to synthetic data quality rather than reconstruction performance alone; (iv) an extensive experimental evaluation comparing LaDIS with established tabular generative baselines in terms of synthetic data fidelity, structural preservation, and computational efficiency, while assessing its downstream utility for multiclass intrusion detection using two complementary classifiers.

\section{Related Work}

Class imbalance is a well-known challenge in ML, as highly skewed datasets can bias models toward majority classes, hindering minority class detection and generalization \cite{Sussumu-Hida26}. Classical strategies such as random oversampling and interpolation-based methods remain widely used due to their simplicity and low computational cost \cite{Azhar22}. However, they do not explicitly model the underlying data distribution and may amplify noise, class overlap, or small disjuncts \cite{Azhar22}. These limitations have motivated the increasing use of DL-based generative models for tabular data synthesis. Conditional Tabular GAN (CTGAN) and Tabular VAE (TVAE) \cite{LeiXu19} are established generative models specifically designed for mixed-type tabular data. CTGAN combines mode-specific normalization, conditional generation, and training-by-sampling to better capture multimodal continuous variables and imbalanced categorical distributions, while TVAE adapts the VAE framework to heterogeneous tabular attributes. Despite their effectiveness, CTGAN is sometimes prone to training instability and mode collapse, whereas TVAE can oversmooth complex distributions \cite{Mallidi25,LeiXu19}. More recently, DMs have emerged as a competitive alternative. Tabular Denoising Diffusion Probabilistic Model (TabDDPM) \cite{Kotelnikov23} separately models numerical and categorical variables through dedicated diffusion processes, although operating directly in the original feature space requires repeated denoising steps and can increase computational cost. Latent diffusion reduces this burden by performing generation in a compact continuous representation. Zhang et al. \cite{Zhang24} introduced TabSyn, which combines a Transformer-based VAE with score-based diffusion in a unified latent space, improving the preservation of feature distributions and inter-column dependencies while enabling efficient sampling. Similarly, Zhu et al. \cite{Zhu25} proposed Differentially Private Tabular Latent Diffusion Model, integrating latent Gaussian diffusion with differentially private AE training to improve the privacy–utility trade-off in synthetic tabular data generation. These contributions establish the effectiveness and versatility of latent diffusion for mixed-type data. However, their primary objectives concern general-purpose tabular synthesis and privacy-preserving generation, respectively, rather than the specific challenges of network intrusion data. In this context, LaDIS extends latent diffusion to attack-specific traffic generation through adaptive selection of compact latent representations and network-aware reconstruction procedures that preserve feature validity and consistency. Its evaluation further emphasizes statistical fidelity, dependency preservation, and the practical utility of synthetic attacks for IDSs.

In IoT IDSs, class imbalance is particularly challenging because minority attacks are not only scarce, but can also overlap with other traffic classes and exhibit complex dependencies among features \cite{Mallidi25}. This has motivated the development of domain-specific augmentation strategies. Dash et al. \cite{Dash20} generated minority samples through nearest-neighbor interpolation, but this approach does not explicitly model the full multivariate traffic distribution and was evaluated on a single dataset. Park et al. \cite{Park23} combined class-specific augmentation with AE-based feature extraction, although the framework remains relatively complex and challenged by overlapping attacks. Rahman et al. \cite{Rahman24} demonstrated the downstream utility of GAN-generated traffic, but assessed synthetic fidelity mainly through marginal comparisons, while Liu et al. \cite{Liu22} combined VAE-based generation with majority-class undersampling, focusing primarily on class balance and classification performance rather than multivariate fidelity. More recently, Xiong et al. \cite{XuanruiXiong26} proposed a VAE-GAN framework based on image-like traffic representations to capture temporal and cross-feature structure, at the cost of a complex multi-stage adversarial pipeline. Wang et al. \cite{ChaolingWang25} introduced a class-conditioned DM with attention and adaptive timesteps, although diffusion is still performed directly in the heterogeneous feature space and its generalization beyond the evaluated scenarios remains unclear.

Existing studies demonstrate the potential of generative augmentation for intrusion detection, but several limitations remain. Many approaches focus on class balancing or downstream classification, with limited evaluation of whether synthetic traffic simultaneously preserves marginal distributions, multivariate geometry, inter-feature dependencies, and attack diversity. Additionally, no generative model has emerged as consistently superior across all evaluation dimensions. These limitations motivated us to propose a latent diffusion framework for IoT attack synthesis evaluated through complementary measures of statistical fidelity, structural preservation, computational efficiency, and downstream utility.

\section{LaDIS Methodology}

This section presents the proposed LaDIS framework. First, the intrusion synthesis problem and the overall architecture are formulated. Then, the attack-specific latent representation learned by the AE is described, followed by the latent diffusion and synthetic generation processes. Finally, the computational complexity of the framework is analyzed.

\subsection{Problem Formulation}
Let $\mathcal{A}$ denote the set of attack types considered for synthetic generation. For each attack $a\in\mathcal{A}$, let
\begin{equation}
    \mathcal{D}^{(a)} = \left\{\mathbf{x}^{(a)}_i\right\}_{i=1}^{N_a}
\end{equation}
denote the corresponding set of real network-traffic observations. Restricting the notation to the variables modeled generatively, each observation contains numerical and categorical information and can be represented as
\begin{equation}
    \mathbf{x}^{(a)}_i = \left(\mathbf{x}^{(a)}_{\mathrm{num},i}, \mathbf{x}^{(a)}_{\mathrm{cat},i}\right) \in \mathbb{R}^{p_a} \times \prod_{j=1}^{q_a} \mathcal{C}^{(a)}_j,
\end{equation}
where $p_a$ and $q_a$ denote the numbers of numerical and categorical variables, respectively, and $\mathcal{C}^{(a)}_j$ is the finite state space associated with the $j$-th categorical variable.

LaDIS follows an attack-specific generative formulation. Rather than fitting a single conditional model across heterogeneous attack classes, an independent model is learned for each $a\in\mathcal{A}$. Consequently, within each $\mathcal{D}^{(a)}$, generation is unconditional and aims to approximate the corresponding attack-specific data distribution. Formally, LaDIS seeks to induce a synthetic distribution $\widehat{p}_a$ such that
\begin{equation}
    \widetilde{\mathbf{x}}^{(a)} \sim \widehat{p}_a, \qquad \widehat{p}_a \approx p_{\mathrm{data}}^{(a)},
\end{equation}
where $p_{\mathrm{data}}^{(a)}$ denotes the unknown joint distribution underlying the real observations of attack $a$. The central idea of LaDIS is to model this heterogeneous distribution indirectly through a continuous latent representation. The framework comprises three successive stages. First, the mixed numerical and categorical observations are represented in a model-compatible space and compressed by an attack-specific mixed-type AE into a lower-dimensional continuous latent space. Second, a Gaussian DM learns the distribution of these latent representations, thus avoiding the need to perform diffusion directly over heterogeneous numerical and categorical variables. Third, synthetic latent samples produced by the learned reverse diffusion process are decoded and mapped back to the network-traffic domain through the corresponding inverse and reconstruction operations. Therefore, for each attack type, LaDIS can be viewed at a high level as the generative mapping
\begin{equation}
    p_{\mathrm{data}}^{(a)} \longrightarrow p_{\mathrm{latent}}^{(a)} \xrightarrow{\;\mathrm{diffusion}\;} \widehat{p}_{\mathrm{latent}}^{(a)} \longrightarrow \widehat{p}_a,
\end{equation}
where $p_{\mathrm{latent}}^{(a)}$ denotes the distribution induced by the learned representation of real attack observations and $\widehat{p}_{\mathrm{latent}}^{(a)}$ the corresponding distribution generated by the latent DM.

The proposed LaDIS framework is publicly available as an open-source Python library.\footnote{\url{https://github.com/estelasc/LaDIS}}

\subsection{Attack-Specific Latent Representation}
Let $\mathcal{T}_a$ denote the training-fitted transformation that maps an observation to the numerical and categorical model representations,
\begin{equation}
    \left(\mathbf{u}^{(a)}_i, \mathbf{c}^{(a)}_i\right) = \mathcal{T}_a \left(\mathbf{x}^{(a)}_i\right),
\end{equation}
where $\mathbf{u}^{(a)}_i\in\mathbb{R}^{p_a}$ contains the transformed numerical variables and
\begin{equation}
    \mathbf{c}^{(a)}_i = \left(c^{(a)}_{i1}, \ldots, c^{(a)}_{iq_a}\right), \qquad c^{(a)}_{ij} \in \left\{0, \ldots, K^{(a)}_j-1\right\},
\end{equation}
contains the zero-based categorical codes, with $K^{(a)}_j$ denoting the cardinality of the $j$-th categorical variable. LaDIS employs a deterministic mixed-type AE to map this heterogeneous representation to a fully continuous latent space. Each categorical variable is first projected through an independent trainable embedding,
\begin{equation}
    \mathbf{e}^{(a)}_{ij} = \mathrm{Emb}^{(a)}_j \left(c^{(a)}_{ij}\right) \in \mathbb{R}^{m^{(a)}_j},
\end{equation}
where $m^{(a)}_j$ denotes the embedding dimension associated with the $j$-th categorical variable. The numerical variables and categorical embeddings are then concatenated and jointly processed by the encoder,
\begin{equation}
    \mathbf{r}^{(a)}_i = E_{\phi_a} \left(\mathbf{u}^{(a)}_i \,\Vert\, \mathbf{e}^{(a)}_{i1} \,\Vert\,\cdots\,\Vert\, \mathbf{e}^{(a)}_{iq_a}\right) \in \mathbb{R}^{d_a},
\end{equation}
where $\Vert$ denotes concatenation and $d_a$ is the attack-specific latent dimensionality. The decoder $D_{\psi_a}$ maps each latent representation back to the mixed model space,
\begin{equation}
    D_{\psi_a} \left(\mathbf{r}^{(a)}_i\right) = \left(\widehat{\mathbf{u}}^{(a)}_i, \left\{\boldsymbol{\ell}^{(a)}_{ij}\right\}_{j=1}^{q_a}\right),
\end{equation}
where $\widehat{\mathbf{u}}^{(a)}_i\in\mathbb{R}^{p_a}$ is the numerical reconstruction and $\boldsymbol{\ell}^{(a)}_{ij}\in\mathbb{R}^{K^{(a)}_j}$ contains the logits associated with categorical variable $j$. Accordingly, the AE is optimized using a mixed-type reconstruction objective,
\begin{equation}
    \mathcal{L}^{(a)}_{\mathrm{AE}} = \lambda_{\mathrm{num}} \mathcal{L}^{(a)}_{\mathrm{num}} + \lambda_{\mathrm{cat}} \mathcal{L}^{(a)}_{\mathrm{cat}},
\end{equation}
with numerical reconstruction loss
\begin{equation}
    \mathcal{L}^{(a)}_{\mathrm{num}} = \frac{1}{B p_a} \sum_{i=1}^{B} \left\| \widehat{\mathbf{u}}^{(a)}_i - \mathbf{u}^{(a)}_i\right\|_2^2,
\end{equation}
and categorical reconstruction loss
\begin{equation}
    \mathcal{L}^{(a)}_{\mathrm{cat}} = \frac{1}{Bq_a} \sum^B_{i=1}\sum_{j=1}^{q_a} \frac{\mathrm{CE} \left(\boldsymbol{\ell}^{(a)}_{ij}, c^{(a)}_{ij}\right)}{\log K^{(a)}_j},
\end{equation}
where $B$ denotes the minibatch size and $\mathrm{CE}(\cdot)$ is the categorical cross-entropy for a single observation. Normalization by $\log K^{(a)}_j$ mitigates disproportionate contributions from categorical variables with larger state spaces. The model checkpoint achieving the lowest validation reconstruction loss is retained.

Once representation learning is complete, the AE parameters are frozen and the encoder is used to obtain the latent representations of the training observations. To provide comparable coordinate scales for the subsequent Gaussian diffusion process, each latent dimension is standardized independently using statistics estimated exclusively from the encoded training partition. Let $N_a^{\mathrm{tr}}$ denote the number of real training observations for attack $a$. Defining
\begin{equation}
    \boldsymbol{\mu}^{(a)}_r = \frac{1}{N_a^{tr}} \sum_{i=1}^{N_a^{tr}} \mathbf{r}^{(a)}_i, \qquad \boldsymbol{\sigma}^{(a)}_r = \sqrt{\frac{1}{N_a^{tr}} \sum_{i=1}^{N_a^{tr}} \left(\mathbf{r}^{(a)}_i - \boldsymbol{\mu}^{(a)}_r\right)^{\odot 2}},
\end{equation}
with $\cdot^{\odot2}$ referring to element-wise squaring of the vector, the standardized clean latent representation is
\begin{equation}
    \mathbf{z}^{(a)}_{0,i} = \mathcal{S}_a \left(\mathbf{r}^{(a)}_i\right) = \frac{\mathbf{r}^{(a)}_i - \boldsymbol{\mu}^{(a)}_r}{\boldsymbol{\sigma}^{(a)}_r},
\end{equation}
with all operations applied coordinate-wise. This transformation centers each training latent coordinate and scales it to unit variance without whitening, rotation, clipping, or additional Gaussianization. The resulting $\mathbf{z}^{(a)}_{0}\in\mathbb{R}^{d_a}$ therefore preserves the dependency structure learned by the AE while defining the continuous space in which the DM operates.

\subsection{Latent Diffusion and Synthetic Generation}
Given $\mathbf{z}^{(a)}_0\in\mathbb{R}^{d_a}$, LaDIS learns their attack-specific distribution using a Gaussian denoising diffusion probabilistic model. For diffusion step $t\in\{1,\ldots,T\}$, let $\beta_t$ denote the noise variance, $\alpha_t=1-\beta_t$, and
\begin{equation}
    \bar{\alpha}_t = \prod_{s=1}^{t}\alpha_s .
\end{equation}
LaDIS uses a cosine noise schedule for $\bar{\alpha}_t$. The forward diffusion process progressively corrupts a clean latent vector according to
\begin{equation}
    q \left(\mathbf{z}^{(a)}_t \mid \mathbf{z}^{(a)}_0\right) = \mathcal{N} \left(\sqrt{\bar{\alpha}_t}\, \mathbf{z}^{(a)}_0, \left(1-\bar{\alpha}_t\right) \mathbf{I}\right),
\end{equation}
or, equivalently,
\begin{equation}
    \mathbf{z}^{(a)}_t = \sqrt{\bar{\alpha}_t}\, \mathbf{z}^{(a)}_0 + \sqrt{1-\bar{\alpha}_t}\, \boldsymbol{\epsilon}, \qquad \boldsymbol{\epsilon} \sim \mathcal{N} \left(\mathbf{0}, \mathbf{I}\right).
\end{equation}
A timestep-conditioned denoiser $f_{\theta_a}$ receives the corrupted latent vector and a sinusoidal representation of $t$, and directly predicts the corresponding clean latent vector,
\begin{equation}
    \widehat{\mathbf{z}}^{(a)}_0 = f_{\theta_a} \left(\mathbf{z}^{(a)}_t, t\right).
\end{equation}
The DM is trained by minimizing the clean latent reconstruction error,
\begin{equation}
    \mathcal{L}^{(a)}_{\mathrm{diff}} = \mathbb{E}_{\mathbf{z}^{(a)}_0,\, t,\, \boldsymbol{\epsilon}} \left[\left\|f_{\theta_a} \left(\mathbf{z}^{(a)}_t,t\right) - \mathbf{z}^{(a)}_0\right\|_2^2\right],
\end{equation}
where $t$ is sampled uniformly across diffusion steps and $\boldsymbol{\epsilon}$ is Gaussian noise.

For generation, the reverse process is constructed from the Gaussian posterior
\begin{equation}
    q \left(\mathbf{z}_{t-1} \mid \mathbf{z}_t, \mathbf{z}_0\right) = \mathcal{N} \left(\widetilde{\boldsymbol{\mu}}_t, \widetilde{\beta}_t \mathbf{I}\right),
\end{equation}
with
\begin{equation}
    \widetilde{\boldsymbol{\mu}}_t = \frac{\sqrt{\bar{\alpha}_{t-1}}\beta_t}{1-\bar{\alpha}_t} \mathbf{z}_0 + \frac{\sqrt{\alpha_t} \left(1-\bar{\alpha}_{t-1}\right)}{1-\bar{\alpha}_t} \mathbf{z}_t,
\end{equation}
and
\begin{equation}
    \widetilde{\beta}_t = \beta_t \frac{1-\bar{\alpha}_{t-1}}{1-\bar{\alpha}_t}.
\end{equation}
Since $\mathbf{z}_0$ is unknown during generation, it is replaced by the denoiser prediction $f_{\theta_a}(\mathbf{z}_t,t)$ in the posterior mean. Starting from
\begin{equation}
    \widetilde{\mathbf{z}}^{(a)}_T \sim \mathcal{N} \left(\mathbf{0}, \mathbf{I}_{d_a}\right),
\end{equation}
ancestral sampling iteratively generates
\begin{equation}
    \widetilde{\mathbf{z}}^{(a)}_{t-1} = \boldsymbol{\mu}_{\theta_a} \left(\widetilde{\mathbf{z}}^{(a)}_t,t\right) + \mathbb{I}_{t>1} \sqrt{\widetilde{\beta}_t}\, \boldsymbol{\epsilon}_t, \qquad \boldsymbol{\epsilon}_t \sim \mathcal{N} \left(\mathbf{0}, \mathbf{I}_{d_a}\right),
\end{equation}
until a synthetic clean latent representation $\widetilde{\mathbf{z}}^{(a)}_0$ is obtained. The generated vector is then returned to the original AE latent coordinates using the training-fitted statistics,
\begin{equation}
    \widetilde{\mathbf{r}}^{(a)} = \boldsymbol{\mu}^{(a)}_r + \boldsymbol{\sigma}^{(a)}_r \odot \widetilde{\mathbf{z}}^{(a)}_0,
\end{equation}
where $\odot$ denotes element-wise multiplication. The frozen decoder produces the numerical model representation and one categorical logit vector per categorical variable,
\begin{equation}
    D_{\psi_a} \left(\widetilde{\mathbf{r}}^{(a)}\right) = \left(\widetilde{\mathbf{u}}^{(a)}, \left\{\boldsymbol{\ell}^{(a)}_j\right\}_{j=1}^{q_a}\right).
\end{equation}
Each categorical state is sampled from its predicted probability distribution,
\begin{equation}
    \widetilde{c}^{(a)}_j \sim \mathrm{Cat} \left(\mathrm{softmax} \left(\boldsymbol{\ell}^{(a)}_j\right)\right).
\end{equation}
Finally, the numerical and categorical model representations are mapped back to the network-traffic domain and the deterministic reconstruction operator $\mathcal{R}_a$ restores the reconstructable structural constraints,
\begin{equation}
    \widetilde{\mathbf{x}}^{(a)} = \mathcal{R}_a \left[\mathcal{T}^{-1}_a \left(\widetilde{\mathbf{u}}^{(a)}, \widetilde{\mathbf{c}}^{(a)}\right)\right].
\end{equation}

\subsection{Computational Complexity Analysis}
Let $P_{\mathrm{AE}}^{(a)}$ denote the number of affine parameters involved in the attack-specific AE and $P_{\mathrm{D}}^{(a)}$ those of the latent denoiser. Since both components are implemented as multilayer perceptrons, their forward and backward computational costs are linear in these quantities. For $N_a^{\mathrm{tr}}$ training observations and $E_a$ AE training epochs, representation learning therefore requires
\begin{equation}
    \mathcal{O} \left(E_a N_a^{\mathrm{tr}} P_{\mathrm{AE}}^{(a)}\right).
\end{equation}
After the AE is frozen, diffusion training is performed directly in the $d_a$-dimensional latent space. If $U_a$ denotes the number of optimization steps and $B_a$ the minibatch size, its computational complexity is
\begin{equation}
    \mathcal{O} \left(U_a B_a P_{\mathrm{D}}^{(a)}\right).
\end{equation}
Note that this cost does not scale multiplicatively with the number of diffusion steps $T$, since a single timestep is sampled for each training observation at every optimization step.

During generation, however, ancestral sampling evaluates the denoiser once at each reverse diffusion step. Generating $M_a$ synthetic observations therefore requires
\begin{equation}
    \mathcal{O} \left(M_a T P_{\mathrm{D}}^{(a)} + M_a P_{\mathrm{dec}}^{(a)}\right),
\end{equation}
where $P_{\mathrm{dec}}^{(a)}$ denotes the computational size of the frozen decoder. The subsequent inverse transformations and deterministic reconstruction are linear in the number of generated observations and modeled variables and are therefore lower-order relative to iterative diffusion sampling.

Consequently, the dominant generation cost is the $T$ repeated denoiser evaluations in the compact latent space. By performing diffusion over $d_a$ continuous latent coordinates rather than directly over the original mixed-type representation, LaDIS confines this iterative generative computation to the learned low-dimensional domain.

\section{Experimental Setup}

This section describes the experimental framework used to evaluate LaDIS against the selected generative baselines. It covers the IoT datasets, attack taxonomy and generation tasks, data preparation and model-specific representations, baseline implementations, model selection and hyperparameter optimization, final training and generation configurations, and the experimental protocol used for synthetic-data evaluation.

\begin{table*}[t]
    \centering
    \caption{Attack taxonomy, data partitions, and synthetic generation targets across the CICIoT23, Bot-IoT, and CETEM datasets.}
    \label{generation_tasks}
    \setlength{\tabcolsep}{3.5pt}
    \resizebox{\textwidth}{!}{%
    \begin{tabular}{@{}llllcccc@{}}
        \hline
        \textbf{Dataset} & \textbf{Attack} & \textbf{Abbreviation} & \textbf{Family} & \textbf{Train} & \textbf{Validation} & \textbf{Test} & \textbf{Synth. Samples} \\
        \hline
        CICIoT23 & Benign traffic & Benign & Benign & - & - & - & - \\
        Bot-IoT & Benign traffic & Benign & Benign & - & - & - & - \\
        CETEM & Benign traffic & Benign & Benign & - & - & - & - \\
        CICIoT23 & Distributed Denial-of-Service User Datagram Protocol Flood & DDoS UDP Flood & Flooding & 393818 & 49227 & 49227 & 125000 \\
        Bot-IoT & Distributed Denial-of-Service User Datagram Protocol & DDoS UDP & Flooding & - & - & - & - \\
        CICIoT23 & Denial-of-Service Transmission Control Protocol Flood & DoS TCP Flood & Flooding & - & - & - & - \\
        Bot-IoT & Denial-of-Service Transmission Control Protocol & DoS TCP & Flooding & 839882 & 104986 & 104985 & 125000 \\
        CICIoT23 & Mirai Greip Flood & Mirai Greip & Botnet & 850000 & 106250 & 106250 & 500000 \\
        CICIoT23 & Recon OS Scan & Recon OS Scan & Recon & 140010 & 17502 & 17501 & 325000 \\
        Bot-IoT & OS Scan & OS Scan & Recon & 142106 & 17763 & 17763 & 325000 \\
        CICIoT23 & Domain Name System Spoofing & DNS Spoofing & Stealthy & 77165 & 9644 & 9646 & 275000 \\
        Bot-IoT & Keylogging & Keylogging & Stealthy & 1178 & 148 & 147 & 270000 \\
        CETEM & Login Poison Virtual Machine & Login P. VM & Stealthy & - & - & - & - \\
        CETEM & Login Virtual Machine & Login VM & Stealthy & - & - & - & - \\
        CICIoT23 & Man-in-the-Middle Address Resolution Protocol Spoofing & MitM ARP Spoofing & Stealthy & 69006 & 8626 & 8626 & 275000 \\
        CETEM & Web Portal Pwned Virtual Machine & Web Portal Pwned VM & Stealthy & - & - & - & - \\
        CETEM & Web Portal Virtual Machine & Web Portal VM & Stealthy & - & - & - & - \\
        \hline
    \end{tabular}
    }
\end{table*}

\subsection{Datasets and Attack-Specific Generation Tasks}
Three IoT network-traffic datasets were considered: CICIoT23 \cite{PintoNeto23}, Bot-IoT \cite{Koroniotis19}, and a non-public dataset provided by the Technological Center for Furniture and Wood of the Region of Murcia (CETEM). The selected traffic types, their unified attack-family taxonomy, and their role in the generation stage are summarized in Table~\ref{generation_tasks}.

Raw packet captures were converted into a common flow-level tabular representation using the same feature-extraction pipeline. Packets were sequentially aggregated into bidirectional internet protocol flows according to endpoint addresses, transport ports, and protocol, with inactive flows closed after 30~s. Each flow was represented through traffic volume, timing, directional, packet length, payload length, header length, inter-arrival time, time to live, transport, and TCP flag descriptors. The complete definition of the resulting 114-column flow table is documented in the public LaDIS repository.\footnote{\url{https://github.com/estelasc/LaDIS/blob/v0.1.1/README.md\#extracted-network-flow-variables}}

Eight attack-specific synthetic generation tasks were defined. For these tasks, Table~\ref{generation_tasks} reports the training, validation, and test partitions, obtained using an approximately 80/10/10 random split of flow-level observations, with identical feature vectors assigned to the same partition, together with the final number of observations generated for each task. Traffic subsets marked with dashes were not used to train attack-specific generators, but were instead retained for the downstream multiclass intrusion detection evaluation.

\subsection{Data Preparation and Model-Space Transformations}
\label{preproc}
A common data preparation pipeline was applied independently to each attack-specific dataset before model fitting. The prepared tables were audited for missing or non-finite values, constant variables, potential identifiers, and high-cardinality fields requiring specific treatment. Observation-specific identifiers and provenance variables were excluded from generative modeling, while recoverable constant metadata were stored separately. Deterministically derived features were removed when their relationships with retained variables could be reliably validated and were reconstructed after generation. The remaining variables were classified as continuous numerical, discrete numerical, or categorical features.

Source and destination ports were tokenized using training-fitted statistics. Up to 50 nonzero ports occurring at least 20 times were retained as individual categorical states, while the remaining values were grouped into well-known, registered, or dynamic/private ranges, with zero and invalid values handled separately. Empirical training distributions were retained to reconstruct concrete port values.

After this common preparation, model-specific representations were constructed. CTGAN and TVAE operated on numerical variables in their original modeling scale and categorical variables in their discrete representation. TabDDPM and LaDIS instead used a transformed mixed-type representation. Strongly right-skewed non-negative numerical variables were first transformed using $\log(1+x)$ and subsequently standardized. Discrete count variables and strongly repeated numerical modes were locally dequantized to remove exact ties, after which a Gaussian quantile transformation fitted exclusively to the training partition was applied to obtain the final continuous numerical model space. Categorical variables were restricted to states observed in the training partition and represented by zero-based integer codes.

All fitted transformations were applied unchanged to the validation and test partitions. After generation, they were inverted in reverse order, including Gaussian-quantile inversion, restoration of repeated modes, inverse numerical scaling, rounding and domain enforcement for discrete variables, recovery of categorical states and concrete ports, and reconstruction of validated deterministic features and constant metadata.

\subsection{Baseline Generative Models}
CTGAN, TVAE \cite{LeiXu19}, and TabDDPM \cite{Kotelnikov23} were used as established tabular generative baselines for comparison with LaDIS. Consistently with the proposed framework, an independent instance of each baseline was trained for every attack-specific generation task. CTGAN and TVAE were implemented using the standalone \texttt{ctgan} package and were deliberately provided with the same modeling variables and preprocessing configuration, so that differences in performance reflected the generative architecture rather than differences in input representation. TabDDPM was implemented from the official source code using its Gaussian--multinomial diffusion formulation and timestep-conditioned multilayer perceptron denoiser. The same definitive set of modeling variables was retained across all baselines, with model-specific representations defined as described in Section~\ref{preproc}.

\subsection{Model Selection and Hyperparameter Optimization}
To ensure controlled model selection, candidate configurations were evaluated under consistent training and validation protocols within each model-specific search stage, using fixed data partitions, random seeds, and stage-specific training budgets. The test partition was excluded from model selection.

For CTGAN, TVAE, and TabDDPM, three controlled configurations were evaluated by modifying one principal architectural or optimization factor with respect to a reference configuration. CTGAN used $G,D=(256,256)$ and $\eta_G=\eta_D=2\times10^{-4}$ as reference, and evaluated either a lower learning rate (LR) of $1\times10^{-4}$ or wider $G,D=(512,512)$ networks. TVAE used encoder and decoder dimensions of $(128,128)$ with $L_2=10^{-5}$ as reference, and compared this configuration against weaker regularization ($L_2=10^{-6}$) and wider $(256,256)$ encoder and decoder networks. TabDDPM used $h=(256,512,512,256)$ and $\eta=5\times10^{-4}$ as reference, and evaluated a lower LR of $2.5\times10^{-4}$ and a wider denoiser $h=(512,1024,1024,512)$. Candidate screening used at most 50000 training observations and 10000 held-out validation observations. CTGAN and TVAE candidates were trained for 50 epochs, whereas TabDDPM candidates were trained for 20000 optimization steps. Synthetic samples were evaluated using the SDMetrics Quality Report. Candidates were ranked primarily by the overall Quality Score, with the percentage of exact matches to training observations used as a tie-breaker.

LaDIS followed a two-stage selection procedure. First, the latent dimension was selected from $d_a\in\{4,8,12,16,20,24,28,32\}$ while keeping the AE and diffusion architectures fixed. For each candidate dimension, a complete AE--diffusion pipeline was trained and evaluated against the same fixed validation reference. Up to 10000 validation observations were used, subject to the number available for the corresponding attack, and an equal number of synthetic observations was generated for each candidate. Candidate dimensions were ranked using the overall SDMetrics Quality Score and its two constituent properties, Column Shapes and Column Pair Trends, and the dimension with the lowest mean rank across these three scores was selected. The AE was then retrained using the selected dimensionality. Second, with the latent representation fixed, four denoiser configurations were evaluated by combining $h\in\{(256,512,512,256),(512,1024,1024,512)\}$ with $\eta\in\{5\times10^{-4},2.5\times10^{-4}\}$. Each candidate was trained for 10000 optimization steps and evaluated against the same fixed validation reference using up to 10000 observations, again limited by the available validation set size, with an equal number of synthetic observations generated for each candidate. Selection followed a deterministic hierarchy, specifically higher Quality Score, lower latent correlation error (CE), smaller deviation of the median generated-to-training latent standard deviation ratio from unity, and fewer model parameters. The selected configuration was retained for definitive synthetic generation.

\subsection{Training and Synthetic Generation Configuration}
After model selection, the definitive baseline models were trained on the training partition using the selected attack-specific configurations. CTGAN and TVAE were trained for 300 epochs with a batch size of 500. TabDDPM was trained for 30000 optimization steps with a batch size of 4096, using 500 diffusion steps, a cosine noise schedule, and $\epsilon$ prediction for the Gaussian component. The TabDDPM checkpoint was selected using a fixed training-subset monitoring score combining noise prediction error at the highest monitored timestep (75\%) and its mean across monitored timesteps (25\%). For LaDIS, the AE corresponding to the selected attack-specific latent dimension used encoder dimensions $(256,128,64)$ and symmetric decoder dimensions. Categorical embedding dimensions were set to $m_j^{(a)}=\min(16,\max(2,\lceil\sqrt{K_j^{(a)}}\rceil))$. The AE reconstruction losses were equally weighted, with $\lambda_{\mathrm{num}}=\lambda_{\mathrm{cat}}=1$. The AE was optimized with AdamW using a LR of $10^{-3}$, weight decay of $10^{-5}$, and a batch size of 4096 for at most 100 epochs, with early stopping after 10 epochs without sufficient validation improvement. The trained AE was subsequently frozen. The selected latent denoiser used 500 diffusion steps with a cosine schedule, a timestep embedding dimension of 128, a batch size of 4096, and AdamW with weight decay $10^{-5}$. Each denoiser candidate had already been trained for 10000 optimization steps on the training latent representation. Therefore, the winning checkpoint from the selection procedure was retained without an additional refit.

Definitive synthetic datasets were generated using ancestral reverse sampling for the diffusion-based models. In LaDIS, categorical variables were sampled from the decoder softmax distributions rather than assigned by maximum probability. The number of synthetic observations generated for each attack-specific task followed the downstream balancing strategy reported in Table~\ref{generation_tasks}. All experiments were GPU-accelerated using an NVIDIA A40.

\subsection{Experimental Protocol}
\label{exp-prot}
Synthetic data quality was evaluated from four complementary perspectives, namely univariate distribution fidelity, multivariate fidelity and geometry, dependency preservation, and global real--synthetic distinguishability. All quantitative metrics were computed on attack-specific feature sets using held-out real test data as reference. Ten evaluation repetitions were performed, each using up to 2500 real test observations and an equal number of synthetic observations per generator, sampled without replacement within each repetition. The same real subset and corresponding synthetic sample positions were reused across generators within each repetition. Results are reported as mean $\pm$ standard deviation over the ten repetitions.

Univariate distribution fidelity was assessed using the normalized Wasserstein distance (NWD) and Kolmogorov--Smirnov (KS) statistic for continuous variables and the total variation distance (TVD) for categorical variables. For each continuous feature, NWD was obtained by dividing the Wasserstein distance between the real and synthetic distributions by the robust range of the held-out real distribution, defined by its $95^{\mathrm{th}}$ and $5^{\mathrm{th}}$ percentiles. NWD was undefined for features with negligible real data robust ranges and was omitted when no valid features remained. Low-cardinality categorical variables were evaluated in their original states, whereas high-cardinality port variables used the training-defined tokenization that preserves frequent ports and groups the remaining values into semantic categories.

Multivariate fidelity and geometry were evaluated using maximum mean discrepancy (MMD), generative precision and recall, and the median nearest-neighbor distance ratio ($N_{\mathrm{med}}$). These metrics were computed in a common feature space fitted exclusively on the real training partition. Continuous variables were scaled using a RobustScaler based on the $5^{\mathrm{th}}$--$95^{\mathrm{th}}$ percentile range and categorical variables were one-hot encoded. MMD was computed using the radial basis function kernel, $k(\mathbf{x},\mathbf{y})=\exp\left(-\gamma\lVert\mathbf{x}-\mathbf{y}\rVert_2^2\right)$, where $\mathbf{x}$ and $\mathbf{y}$ denote arbitrary observations in the transformed feature space and $\gamma$ controls the kernel bandwidth. A single $\gamma$ was estimated from up to 2000 transformed real training observations using
\begin{equation}
    \gamma=\frac{1}{2\cdot\mathrm{median}\left(d_{ij}^{2}\right)},
\end{equation}
where $d_{ij}=\lVert\mathbf{x}_i-\mathbf{x}_j\rVert_2$ is the Euclidean distance between the $i$-th and $j$-th transformed real training observations, with $i\neq j$. Zero pairwise distances were excluded. The same $\gamma$ was used for all generators and evaluation repetitions. The reported MMD corresponds to the square root of the biased squared MMD estimate. Generative precision and recall were computed in the same transformed space using $k=3$ nearest-neighbor manifold radii. $N_{\mathrm{med}}$ was defined as the median synthetic-to-real nearest-neighbor distance divided by the median real-to-real nearest-neighbor distance in the corresponding held-out real subset, excluding self-distances. Values of $N_{\mathrm{med}}$ close to 1 reveal a real-like local neighborhood scale.

Dependency preservation was assessed using the median absolute CE (ACE), maximum CE, and mutual information (MI) error. CEs were obtained from the element-wise absolute differences between the real and synthetic Spearman correlation matrices. The median and maximum errors over the unique off-diagonal feature pairs were reported as median ACE and maximum CE. For MI analysis, continuous variables were discretized into up to 10 quantile-based bins whose boundaries were estimated exclusively from the real training partition, while categorical variables retained their discrete states. For two variables $X$ and $Y$, MI was normalized as
\begin{equation}
    \mathrm{MI_{norm}}(X,Y)=\frac{2\cdot I(X;Y)}{H(X)+H(Y)},
\end{equation}
where $I(X;Y)$ denotes their MI and $H(X)$ and $H(Y)$ their respective entropies. MI error was computed as the mean absolute difference between the real and synthetic normalized MI matrices over the unique feature pairs.

Global real--synthetic distinguishability was evaluated using a classifier two-sample test (C2ST), reported as the area under the receiver operating characteristic curve (ROC-AUC). The classifier consisted of two hidden layers with 128 and 64 rectified linear unit a followed by a binary output unit. Equal-sized real and synthetic subsets were evaluated using five-fold stratified cross-validation with shuffling. For each fold, the classifier was trained for 75 full-batch epochs using AdamW with LR $\eta=10^{-3}$ and weight decay $10^{-4}$, minimizing binary cross-entropy with logits. ROC-AUC was computed from the pooled out-of-fold probabilities. The same fold assignments and fold-specific initialization seeds were reused across generators within each evaluation repetition. A C2ST ROC-AUC close to 0.5 indicates low real--synthetic distinguishability.

For qualitative dependency analysis, absolute pairwise differences between the real and synthetic Spearman correlation matrices were computed for each generator and evaluation repetition. For each attack and generator, the resulting CE matrices were first averaged element-wise over the ten repetitions. The four generator-specific mean CE matrices were then averaged to obtain a common aggregate CE matrix used only for feature selection. For each feature, its mean CE with all remaining features was computed, and the 15 features with the largest values were selected. Feature selection was performed independently for each attack. Within a given attack, the same features, ordering, and color scale were used for all generators.

For manifold analysis, hierarchical two-dimensional Uniform Manifold Approximation and Projection (UMAP) embeddings were constructed using real observations as the reference space. Up to 600 real observations were sampled per attack and, when available, an equal number of LaDIS-generated observations were considered. Only numerical features shared across all considered cohorts were retained. Missing values were imputed using medians estimated from the real observations. Variables that were constant within the corresponding real reference subset were removed, and the remaining variables were robustly scaled using the $5^{\mathrm{th}}$ and $95^{\mathrm{th}}$ percentiles. Two embedding levels were considered. A global UMAP was first fitted on the real observations to represent the broad attack-family structure. Separate UMAP models were subsequently fitted within each attack family to represent individual attack structure. In both cases, two components, Euclidean distance, 30 neighbors, and a minimum distance of 0.10 were used. All preprocessing and UMAP models were fitted on real data, with synthetic observations subsequently projected into the fixed real-data embeddings.

To assess the downstream utility of LaDIS-generated traffic, CICIoT23, Bot-IoT, and CETEM observations were grouped into five broad traffic classes. A fixed imbalanced real-data design was constructed with 200000, 150000, 100000, 60000, and 20000 training observations for the Benign, Flooding, Botnet, Recon, and Stealthy classes, respectively. The corresponding validation and test sets contained 25000, 18750, 12500, 7500, and 2500 real observations per class, preserving the original 80/10/10 train/validation/test partitions and the relative contribution of individual attacks within each family. Three training regimes were compared, with $\lambda\in\{0,0.5,1\}$ denoting the fraction of the class-size gap filled by synthetic observations: real-only training, partial augmentation, and complete balancing, respectively. Validation and test sets remained entirely real and unchanged. All data-dependent preprocessing was fitted independently within each repetition using only the corresponding real training subset and then applied unchanged to synthetic, validation, and test observations. Real and synthetic observations used the same IDS feature schema, excluding identifiers and provenance variables.

Downstream classification was evaluated using TabNet \cite{Arik21} and XGBoost \cite{Chen16} to assess whether the effect of synthetic augmentation was consistent across different tabular learning paradigms. TabNet provides a neural attention-based approach to tabular learning, whereas XGBoost provides a complementary gradient-boosted tree-based model. TabNet used $n_d=n_a=16$, five decision steps, $\gamma=1.3$, categorical embedding dimension 3, sparse regularization $\lambda_{\mathrm{sparse}}=10^{-3}$, sparsemax masks, Adam optimization with LR $2\times10^{-2}$ and weight decay $10^{-5}$, and a StepLR scheduler with step size 10 and factor 0.9. Training used batches of 8192 observations, virtual batches of 512, a maximum of 100 epochs, and early stopping with patience 20 on the real validation set. XGBoost used the multiclass soft-probability objective with 2000 trees, LR 0.05, maximum depth 8, minimum child weight 1, subsampling and column subsampling rates of 0.8, $L_2$ regularization of 1, no $L_1$ regularization, histogram-based tree construction, and early stopping after 50 rounds without validation improvement, using multiclass log-loss as the evaluation metric. Both classifiers used identical sampled subsets and preprocessing, were GPU-accelerated, and were evaluated over five repetitions using independent random seeds. Final predictions were obtained by the standard maximum-probability decision rule. Performance was evaluated on the untouched real test set using balanced accuracy, macro F1-score, and class-wise precision, recall, and F1-score. To further assess the effect of augmentation under class imbalance, Benign, Flooding, and Botnet were grouped as high-volume families, while Recon and Stealthy were grouped as low-volume families, reflecting their substantially smaller real training set representation. Group-level F1-scores were computed as unweighted averages of the corresponding class-wise F1-scores.

\begin{table*}[t]
    \centering
    \caption{Sensitivity of the proposed LaDIS to the AE latent dimensionality. Bold values indicate the best result for each metric.}
    \label{ldim}
    \setlength{\tabcolsep}{3.5pt}
    \renewcommand{\arraystretch}{1.05}
    \scriptsize
    \resizebox{\textwidth}{!}{%
    \begin{tabular}{llcccccccc}
        \toprule
        \textbf{Attack} & \textbf{Metric} & $\mathbf{d=4}$ & $\mathbf{d=8}$ & $\mathbf{d=12}$ & $\mathbf{d=16}$ & $\mathbf{d=20}$ & $\mathbf{d=24}$ & $\mathbf{d=28}$ & $\mathbf{d=32}$ \\
        \midrule

        \multirow{2}{*}{DoS TCP}
        & AE RMSE ($\downarrow$) & 0.42 & 0.34 & 0.29 & 0.25 & 0.23 & 0.20 & 0.14 & \textbf{0.12} \\
        & Generative mean rank ($\downarrow$) & 6.00 & \textbf{2.00} & 2.67 & 4.33 & 7.33 & 6.67 & 2.33 & 4.67 \\
        \midrule

        \multirow{3}{*}{Mirai Greip}
        & AE RMSE ($\downarrow$) & 0.92 & 0.86 & 0.81 & 0.78 & 0.74 & 0.69 & 0.67 & \textbf{0.64} \\
        & Generative mean rank ($\downarrow$) & 5.67 & \textbf{2.00} & 4.33 & 4.67 & 7.67 & 4.33 & 3.33 & 4.00 \\
        \midrule

        \multirow{3}{*}{Recon OS Scan}
        & AE RMSE ($\downarrow$) & 0.60 & 0.56 & 0.52 & 0.49 & 0.46 & 0.43 & 0.40 & \textbf{0.37} \\
        & Generative mean rank ($\downarrow$) & 2.33 & \textbf{2.00} & 4.67 & 4.33 & 4.33 & 5.33 & 5.67 & 7.33 \\
        \midrule

        \multirow{3}{*}{MitM ARP Spoofing}
        & AE RMSE ($\downarrow$) & 0.55 & 0.51 & 0.47 & 0.43 & 0.40 & 0.36 & 0.33 & \textbf{0.31} \\
        & Generative mean rank ($\downarrow$) & 5.67 & 6.33 & 3.67 & 4.67 & 3.33 & 5.00 & \textbf{2.67} & 4.67 \\

        \bottomrule
    \end{tabular}%
    }
\end{table*}

\begin{table*}[t]
    \centering
    \caption{Quantitative comparison of synthetic data fidelity across representative attacks. Values are reported as mean $\pm$ standard deviation over ten evaluations using held-out real test and synthetic subsamples. Best results for each metric and attack are shown in bold. A dash indicates an undefined metric.\vspace{1mm} }
    \label{quantitative}

    \setlength{\tabcolsep}{3.5pt}
    \renewcommand{\arraystretch}{1.08}
    \footnotesize

    \begin{tabular}{lcccccccc}
        \toprule
        & \multicolumn{4}{c}{\textbf{DoS TCP}}
        & \multicolumn{4}{c}{\textbf{Mirai Greip}} \\
        \cmidrule(lr){2-5}
        \cmidrule(lr){6-9}

        \textbf{Metric}
        & CTGAN & TVAE & TabDDPM & LaDIS & CTGAN & TVAE & TabDDPM & LaDIS \\
        \midrule

        Median NWD ($\downarrow$) & 0.04 $\pm$ 0.01 & 0.03 $\pm$ 0.01 & \textbf{0.02 $\pm$ 0.00} & \textbf{0.02 $\pm$ 0.00} & - & - & - & - \\

        Median KS ($\downarrow$) & 0.06 $\pm$ 0.01 & 0.09 $\pm$ 0.01 & \textbf{0.03 $\pm$ 0.00} & 0.06 $\pm$ 0.01 & 0.43 $\pm$ 0.01 & \textbf{0.00 $\pm$ 0.00} & \textbf{0.00 $\pm$ 0.00} & \textbf{0.00 $\pm$ 0.00} \\

        Median TVD ($\downarrow$) & 0.05 $\pm$ 0.01 & 0.02 $\pm$ 0.01 & \textbf{0.01 $\pm$ 0.01} & 0.02 $\pm$ 0.01 & 0.40 $\pm$ 0.00 & 0.26 $\pm$ 0.01 & \textbf{0.01 $\pm$ 0.00} & 0.02 $\pm$ 0.00 \\

        MMD ($\downarrow$) & 0.05 $\pm$ 0.01 & 0.07 $\pm$ 0.00 & \textbf{0.03 $\pm$ 0.00} & 0.07 $\pm$ 0.01 & 1.07 $\pm$ 0.00 & 0.59 $\pm$ 0.01 & 0.13 $\pm$ 0.00 & \textbf{0.02 $\pm$ 0.00} \\

        Precision ($\uparrow$) & \textbf{0.59 $\pm$ 0.43} & 0.53 $\pm$ 0.48 & 0.54 $\pm$ 0.49 & 0.58 $\pm$ 0.45 & 0.76 $\pm$ 0.09 & 0.61 $\pm$ 0.21 & 0.85 $\pm$ 0.08 & \textbf{1.00 $\pm$ 0.00} \\

        Recall ($\uparrow$) & 0.97 $\pm$ 0.01 & 0.90 $\pm$ 0.04 & \textbf{0.99 $\pm$ 0.00} & 0.40 $\pm$ 0.03 & \textbf{1.00 $\pm$ 0.00} & \textbf{1.00 $\pm$ 0.00} & \textbf{1.00 $\pm$ 0.00} & \textbf{1.00 $\pm$ 0.00} \\

        $N_{\mathrm{med}}$ Ratio ($\approx 1$) & 1.01 $\pm$ 0.00 & 1.15 $\pm$ 0.01 & 1.10 $\pm$ 0.00 & \textbf{1.00 $\pm$ 0.00} & 0.00 $\pm$ 0.00 & 0.82 $\pm$ 0.15 & \textbf{1.00 $\pm$ 0.00} & \textbf{1.00 $\pm$ 0.00} \\

        Median ACE ($\downarrow$) & \textbf{0.01 $\pm$ 0.00} & 0.03 $\pm$ 0.00 & \textbf{0.01 $\pm$ 0.00} & \textbf{0.01 $\pm$ 0.00} & 0.23 $\pm$ 0.10 & 0.31 $\pm$ 0.16 & 0.28 $\pm$ 0.13 & \textbf{0.10 $\pm$ 0.04} \\

        Maximum CE ($\downarrow$) & 0.95 $\pm$ 0.16 & 0.97 $\pm$ 0.10 & 0.85 $\pm$ 0.32 & \textbf{0.45 $\pm$ 0.38} & \textbf{1.06 $\pm$ 0.09} & 1.07 $\pm$ 0.06 & 1.57 $\pm$ 0.19 & 1.24 $\pm$ 0.20 \\

        MI Error ($\downarrow$) & \textbf{0.03 $\pm$ 0.00} & 0.07 $\pm$ 0.00 & 0.05 $\pm$ 0.00 & \textbf{0.03 $\pm$ 0.00} & 0.04 $\pm$ 0.00 & 0.04 $\pm$ 0.00 & 0.04 $\pm$ 0.00 & \textbf{0.02 $\pm$ 0.00} \\

        C2ST ROC-AUC ($\approx 0.50$) & \textbf{0.56 $\pm$ 0.01} & 0.79 $\pm$ 0.01 & 0.72 $\pm$ 0.02 & 0.68 $\pm$ 0.01 & 0.99 $\pm$ 0.00 & 0.76 $\pm$ 0.01 & 0.58 $\pm$ 0.00 & \textbf{0.50 $\pm$ 0.01} \\

        \midrule
        \midrule

        & \multicolumn{4}{c}{\textbf{Recon OS Scan}}
        & \multicolumn{4}{c}{\textbf{MitM ARP Spoofing}} \\
        \cmidrule(lr){2-5}
        \cmidrule(lr){6-9}

        \textbf{Metric}
        & CTGAN & TVAE & TabDDPM & LaDIS & CTGAN & TVAE & TabDDPM & LaDIS \\
        \midrule

        Median NWD ($\downarrow$) & 0.16 $\pm$ 0.02 & 0.19 $\pm$ 0.04 & 11.99 $\pm$ 1.94 & \textbf{0.10 $\pm$ 0.01} & 0.12 $\pm$ 0.02 & 0.08 $\pm$ 0.01 & 2.63 $\pm$ 0.18 & \textbf{0.07 $\pm$ 0.01} \\

        Median KS ($\downarrow$) & 0.14 $\pm$ 0.01 & 0.17 $\pm$ 0.01 & 0.33 $\pm$ 0.00 & \textbf{0.04 $\pm$ 0.01} & 0.23 $\pm$ 0.01 & 0.19 $\pm$ 0.00 & 0.27 $\pm$ 0.00 & \textbf{0.06 $\pm$ 0.00} \\

        Median TVD ($\downarrow$) & 0.21 $\pm$ 0.01 & 0.21 $\pm$ 0.01 & \textbf{0.09 $\pm$ 0.01} & 0.10 $\pm$ 0.01 & 0.34 $\pm$ 0.01 & 0.08 $\pm$ 0.01 & 0.12 $\pm$ 0.01 & \textbf{0.05 $\pm$ 0.01} \\

        MMD ($\downarrow$) & 0.12 $\pm$ 0.01 & 0.11 $\pm$ 0.00 & 0.69 $\pm$ 0.01 & \textbf{0.04 $\pm$ 0.00} & 0.16 $\pm$ 0.01 & 0.12 $\pm$ 0.01 & 0.70 $\pm$ 0.00 & \textbf{0.06 $\pm$ 0.01} \\

        Precision ($\uparrow$) & 0.34 $\pm$ 0.07 & 0.20 $\pm$ 0.08 & 0.20 $\pm$ 0.05 & \textbf{0.44 $\pm$ 0.01} & \textbf{0.31 $\pm$ 0.02} & 0.25 $\pm$ 0.04 & 0.03 $\pm$ 0.01 & 0.26 $\pm$ 0.02 \\

        Recall ($\uparrow$) & 0.83 $\pm$ 0.03 & 0.94 $\pm$ 0.01 & \textbf{0.99 $\pm$ 0.01} & 0.87 $\pm$ 0.02 & 0.88 $\pm$ 0.02 & 0.93 $\pm$ 0.02 & \textbf{0.98 $\pm$ 0.01} & 0.92 $\pm$ 0.01 \\

        $N_{\mathrm{med}}$ Ratio ($\approx 1$) & 1.49 $\pm$ 0.03  & 1.74 $\pm$ 0.02 & 26.26 $\pm$ 12.38 & \textbf{1.02 $\pm$ 0.00} & 1.86 $\pm$ 0.06 & 2.16 $\pm$ 0.06 & 73.48 $\pm$ 13.25 & \textbf{1.08 $\pm$ 0.04} \\

        Median ACE ($\downarrow$) & 0.16 $\pm$ 0.01 & 0.12 $\pm$ 0.00 & 0.28 $\pm$ 0.00 & \textbf{0.04 $\pm$ 0.00} & 0.13 $\pm$ 0.01 & 0.13 $\pm$ 0.01 & 0.25 $\pm$ 0.00 & \textbf{0.05 $\pm$ 0.01} \\

        Maximum CE ($\downarrow$) & 1.00 $\pm$ 0.01 & \textbf{1.00 $\pm$ 0.00} & 1.45 $\pm$ 0.01 & \textbf{1.00 $\pm$ 0.00} & 0.74 $\pm$ 0.04 & 0.81 $\pm$ 0.02 & 1.66 $\pm$ 0.01 & \textbf{0.59 $\pm$ 0.04} \\

        MI Error ($\downarrow$) & 0.12 $\pm$ 0.00 & 0.13 $\pm$ 0.00 & 0.20 $\pm$ 0.00 & \textbf{0.04 $\pm$ 0.00} & 0.10 $\pm$ 0.00 & 0.11 $\pm$ 0.00 & 0.15 $\pm$ 0.00 & \textbf{0.05 $\pm$ 0.00} \\

        C2ST ROC-AUC ($\approx 0.50$) & 0.77 $\pm$ 0.01 & 0.83 $\pm$ 0.01 & 1.00 $\pm$ 0.00 & \textbf{0.67 $\pm$ 0.01} & 0.84 $\pm$ 0.01 & 0.80 $\pm$ 0.01 & 0.95 $\pm$ 0.01 & \textbf{0.70 $\pm$ 0.01} \\

        \bottomrule
    \end{tabular}
\end{table*}


\section{Results}

This section presents the results obtained under the experimental protocol of Section~\ref{exp-prot}. Detailed quality analyses are reported for one representative attack per malicious traffic family: DoS TCP for Flooding, Mirai Greip for Botnet, Recon OS Scan for Recon, and MitM ARP Spoofing for Stealthy. Computational efficiency and downstream IDS performance are also evaluated.

Since LaDIS relies on the representation learned by its AE, the effect of latent dimensionality was evaluated for $d_a\in\{4,8,12,16,20,24,28,32\}$ while keeping the remaining configuration fixed. As shown in Table~\ref{ldim}, increasing $d_a$ consistently reduced validation root mean square error, indicating improved reconstruction. However, better reconstruction did not necessarily translate into better generation. For each attack, candidate dimensions were ranked separately by overall Quality Score, Column Shapes, and Column Pair Trends, and the three ranks were averaged to obtain the generative mean rank, with lower values indicating better performance. This criterion selected $d_a=8$ for DoS TCP, Mirai Greip, and Recon OS Scan, and $d_a=28$ for MitM ARP Spoofing.

\begin{table*}[t]
    \centering
    \caption{Computational efficiency of CTGAN, TVAE, TabDDPM, and LaDIS across representative attacks. Training times report the fitting cost of the selected model configurations, excluding additional hyperparameter search candidates. Sampling times correspond to 10000 observations, with LaDIS sampling comprising latent diffusion and AE decoding.}
    \label{efficiency}

    \setlength{\tabcolsep}{4pt}
    \renewcommand{\arraystretch}{1.08}
    \small

    \begin{tabular}{lcccccccc}
        \toprule
        & \multicolumn{2}{c}{\textbf{DoS TCP}}
        & \multicolumn{2}{c}{\textbf{Mirai Greip}}
        & \multicolumn{2}{c}{\textbf{Recon OS Scan}}
        & \multicolumn{2}{c}{\textbf{MitM ARP Spoofing}} \\
        
        \cmidrule(lr){2-3}
        \cmidrule(lr){4-5}
        \cmidrule(lr){6-7}
        \cmidrule(lr){8-9}

        \textbf{Model} & Train (min) & Sample (s) & Train (min) & Sample (s) & Train (min) & Sample (s) & Train (min) & Sample (s) \\
        \midrule

        CTGAN & 489.88 & 2.51 $\pm$ 0.09 & 551.89 & 2.64 $\pm$ 0.04 & 99.90 & 2.77 $\pm$ 0.01 & 46.10 & 2.76 $\pm$ 0.02 \\

        TVAE & 341.04 & 2.39 $\pm$ 0.01 & 455.04 & 2.51 $\pm$ 0.05 & 82.24 & 2.52 $\pm$ 0.10 & 40.02 & 2.55 $\pm$ 0.15 \\

        TabDDPM & 9.99 & 14.97 $\pm$ 0.08 & 9.03 & 10.78 $\pm$ 0.13 & 8.27 & 13.88 $\pm$ 0.52 & 7.83 & 14.57 $\pm$ 0.22 \\

        LaDIS AE & 11.30 & 0.02 $\pm$ 0.00 & 9.02 & 0.01 $\pm$ 0.00 & 1.49 & 0.01 $\pm$ 0.00 & 0.72 & 0.08 $\pm$ 0.00 \\

        LaDIS Diffusion & 3.21 & 3.11 $\pm$ 0.01 & 2.14 & 2.63 $\pm$ 0.07 & 2.03 & 3.05 $\pm$ 0.03 & 2.23 & 3.12 $\pm$ 0.07 \\

        \bottomrule
    \end{tabular}
\end{table*}

Table~\ref{quantitative} compares LaDIS with CTGAN, TVAE, and TabDDPM across the four representative malicious traffic families. Generally, LaDIS provided a competitive balance between marginal fidelity, multivariate structure, dependency preservation, and real--synthetic indistinguishability, although its relative performance varied across attacks. For DoS TCP, TabDDPM achieved the strongest marginal fidelity and lowest MMD, together with the highest recall. CTGAN obtained the highest precision and the C2ST ROC-AUC closest to 0.5. LaDIS, however, preserved the local neighborhood scale particularly well, with $N_{\mathrm{med}}=1.00$, and achieved the lowest maximum CE and a low MI error. Its main limitation was recall, which decreased to 0.40, indicating incomplete coverage of the real distribution despite good fidelity within the generated regions. For DoS TCP, generative precision exhibited substantial variability across evaluation repetitions, with highly correlated fluctuations across all models. This suggests sensitivity to the sampled reference geometry, and small differences in mean precision should therefore be interpreted cautiously. For Mirai Greip, LaDIS showed the clearest overall advantage, achieving an MMD of 0.02, precision and recall of 1.00, $N_{\mathrm{med}}=1.00$, the lowest median ACE and MI error, and a C2ST ROC-AUC of 0.50. TabDDPM slightly outperformed LaDIS in TVD and also reproduced the local neighborhood scale accurately, but showed larger dependency errors and greater real--synthetic distinguishability. In contrast, the $N_{\mathrm{med}}=0.00$ obtained by CTGAN indicates an abnormally collapsed nearest-neighbor scale, which, together with its high MMD and C2ST ROC-AUC, did not correspond to faithful reproduction of the overall distribution. LaDIS also performed strongly for Recon OS Scan and MitM ARP Spoofing. For Recon OS Scan, it obtained the best or near-best results across most fidelity and dependency metrics, including the lowest NWD, KS, MMD, median ACE, MI error, and C2ST ROC-AUC, while maintaining $N_{\mathrm{med}}$ close to one. TabDDPM achieved the highest recall, but its very large NWD and $N_{\mathrm{med}}$ values revealed substantial marginal and local-geometry distortion. The same trade-off was more pronounced for MitM ARP Spoofing, where LaDIS achieved the best values for most distributional and dependency metrics and $N_{\mathrm{med}}=1.08$, whereas TabDDPM again maximized recall at the expense of strongly distorted neighborhood geometry and higher distinguishability. For all attacks, these results show that high manifold coverage alone does not guarantee high synthetic-data fidelity. TabDDPM frequently attained high recall while distorting marginal distributions or local geometry, whereas LaDIS generally provided a more balanced preservation of distributional, geometric, and dependency structure.

\begin{figure}
    \centering
    \includegraphics[width=.9\linewidth]{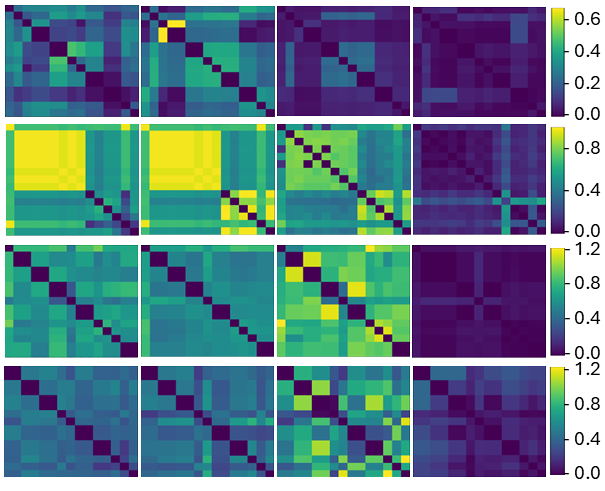}
    \caption{Heatmaps of the absolute differences between the Spearman correlation matrices of real and synthetic data for CTGAN, TVAE, TabDDPM, and LaDIS (left to right), evaluated on DoS TCP, Mirai Greip, Recon OS Scan, and MitM ARP Spoofing (top to bottom).}
    \label{spearman}
\end{figure}

To complement the quantitative analysis, Fig.~\ref{spearman} shows the pairwise CEs across the representative attacks. For DoS TCP, the differences between generators are relatively moderate, although LaDIS exhibits fewer pronounced error regions. For Mirai Greip, CTGAN and TVAE show broad dependency mismatches, while TabDDPM reduces these discrepancies but retains several marked interactions. LaDIS shows substantially lower errors across most feature pairs, with only a few localized discrepancies. The contrast is particularly clear for Recon OS Scan and MitM ARP Spoofing, where TabDDPM exhibits several high-error regions, while LaDIS remains predominantly close to zero. These visual patterns support the quantitative CE results and indicate more consistent preservation of the real correlation structure by LaDIS.

Computational efficiency was evaluated in terms of training and sampling time, as reported in Table~\ref{efficiency}. For LaDIS, the AE and latent diffusion stages are reported separately to characterize their computational contributions, with the total training cost corresponding to their sum. For the larger DoS TCP and Mirai Greip training sets, the AE accounted for most of this cost, requiring 11.30 and 9.02 min, respectively, compared with 3.21 and 2.14 min for latent diffusion training. For the smaller Recon OS Scan and MitM ARP Spoofing datasets, AE training decreased to 1.49 and 0.72 min, yielding total LaDIS training times of 3.52 and 2.95 min. LaDIS training was slower than TabDDPM for DoS TCP and Mirai Greip, but considerably faster for Recon OS Scan and MitM ARP Spoofing, while CTGAN and TVAE required considerably longer training times in all cases. For sampling, the computational cost of LaDIS was almost entirely determined by the latent reverse-diffusion process, requiring 2.63--3.12 s for 10000 samples, whereas AE decoding required only 0.01--0.08 s. Although CTGAN and TVAE provided slightly faster direct sampling, TabDDPM required 10.78--14.97 s for the same workload. Thus, performing diffusion in the latent space clearly reduced sampling cost relative to direct-space diffusion, while AE decoding introduced negligible additional overhead.

\begin{figure}
    \centering
    \includegraphics[width=.85\linewidth]{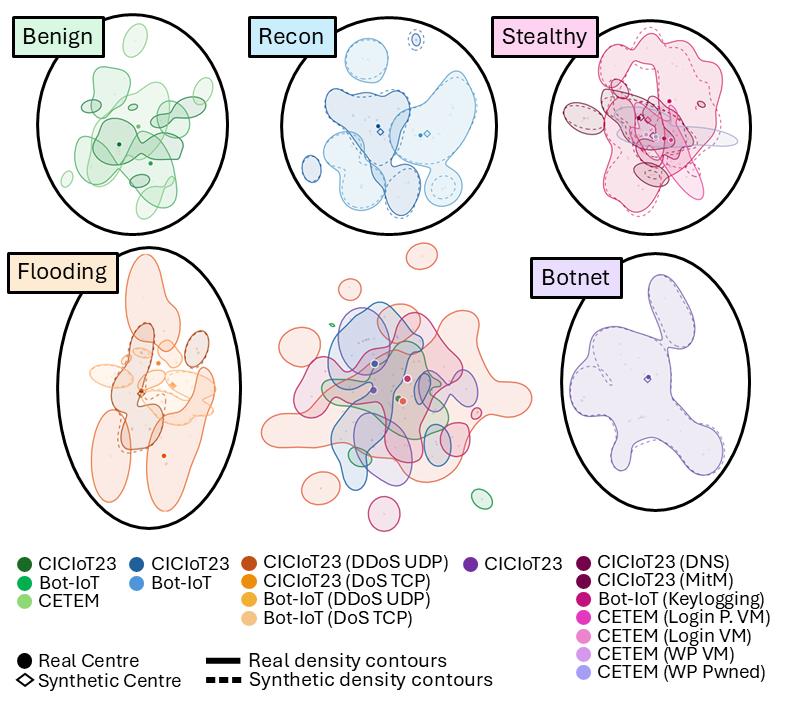}
    \caption{Hierarchical UMAP representation of real and LaDIS-generated traffic, including the global organization of the five traffic families (central panel) and family-specific attack distributions (surrounding panels). Solid and dashed contours enclose approximately 80\% of the real and synthetic probability mass, respectively, estimated by kernel density estimation. Filled circles and open diamonds indicate the corresponding real and synthetic median centers.}
    \label{embedding}
\end{figure}

\begin{figure*}[t]
    \centering
    \begin{tikzpicture}
    \begin{groupplot}[
        group style={
            group size=2 by 1,
            horizontal sep=1.4cm},
        width=0.46\textwidth,
        height=5.3cm,
        xmin=0.60,
        xmax=1.00,
        xtick={0.6,0.7,0.8,0.9,1.0},
        xlabel={Score},
        ytick={1,2,3,4},
        yticklabels={
            Low-volume F1,
            High-volume F1,
            Macro F1,
            Balanced Accuracy},
        yticklabel style={
            font=\footnotesize,
            align=right},
        tick label style={font=\footnotesize},
        label style={font=\small},
        title style={font=\small\bfseries},
        grid=both,
        xmajorgrids=true,
        ymajorgrids=false,
        axis line style={black!60},
        error bars/.cd,
    ]
    
    \nextgroupplot[title={(a) TabNet}]
    \addplot+[
        only marks,
        mark=o,
        mark size=2.6pt,
        error bars/.cd,
            x dir=both,
            x explicit]
    coordinates {
        (0.73,1) +- (0.02,0)
        (0.97,2) +- (0.00,0)
        (0.87,3) +- (0.01,0)
        (0.86,4) +- (0.01,0)};
    \addplot+[
        only marks,
        mark=square*,
        mark size=2.6pt,
        error bars/.cd,
            x dir=both,
            x explicit]
    coordinates {
        (0.84,1) +- (0.01,0)
        (0.98,2) +- (0.00,0)
        (0.93,3) +- (0.01,0)
        (0.94,4) +- (0.00,0)};
    \addplot+[
        only marks,
        mark=triangle*,
        mark size=3pt,
        error bars/.cd,
            x dir=both,
            x explicit]
    coordinates {
        (0.67,1) +- (0.02,0)
        (0.95,2) +- (0.01,0)
        (0.84,3) +- (0.01,0)
        (0.88,4) +- (0.00,0)};
    
    \nextgroupplot[
        title={(b) XGBoost},
        yticklabels={}]
    \addplot+[
        only marks,
        mark=o,
        mark size=2.6pt,
        error bars/.cd,
            x dir=both,
            x explicit]
    coordinates {
        (0.82,1) +- (0.00,0)
        (0.97,2) +- (0.00,0)
        (0.91,3) +- (0.00,0)
        (0.92,4) +- (0.00,0)};
    \addplot+[
        only marks,
        mark=square*,
        mark size=2.6pt,
        error bars/.cd,
            x dir=both,
            x explicit]
    coordinates {
        (0.88,1) +- (0.01,0)
        (0.98,2) +- (0.00,0)
        (0.95,3) +- (0.00,0)
        (0.96,4) +- (0.01,0)};
    \addplot+[
        only marks,
        mark=triangle*,
        mark size=3pt,
        error bars/.cd,
            x dir=both,
            x explicit]
    coordinates {
        (0.82,1) +- (0.00,0)
        (0.97,2) +- (0.00,0)
        (0.91,3) +- (0.00,0)
        (0.92,4) +- (0.00,0)};
    \end{groupplot}
    
    \node[
        anchor=north,
        font=\small
    ] at ($(group c1r1.south)!0.5!(group c2r1.south)+(0,-1.1cm)$)
    {
        \begin{tikzpicture}[baseline]
            \node at (0,0) {
                \tikz{\draw plot[only marks, mark=o, mark size=2.6pt]
                    coordinates {(0,0)};}
                \hspace{0.05cm} Real-only training ($\lambda=0$)
                \hspace{0.5cm}
                \tikz{\draw plot[only marks, mark=square*, mark size=2.6pt]
                    coordinates {(0,0)};}
                \hspace{0.05cm} Partial augmentation ($\lambda=0.5$)
                \hspace{0.5cm}
                \tikz{\draw plot[only marks, mark=triangle*, mark size=3pt]
                    coordinates {(0,0)};}
                \hspace{0.05cm} Complete balancing ($\lambda=1$)
            };
        \end{tikzpicture}};
    \end{tikzpicture}
    \caption{Downstream IDS performance under real-only training, partial LaDIS augmentation, and complete LaDIS balancing for (a) TabNet and (b) XGBoost. Points indicate mean performance, and horizontal error bars indicate the standard deviation across the five repetitions.}
    \label{ids_results}
\end{figure*}

Figure~\ref{embedding} provides a hierarchical view of the real and LaDIS-generated traffic manifolds, combining the global organization of the five traffic families with family-specific attack distributions. In general, synthetic centers lie close to their real counterparts and the corresponding density contours show substantial overlap, indicating that LaDIS reproduces the main regions occupied by real traffic. This agreement is particularly informative for Flooding, Recon, and Stealthy, where several attack types coexist within the same family. LaDIS preserves distinct attack-specific regions rather than collapsing them into a common synthetic distribution, including differences between attacks originating from CICIoT23 and Bot-IoT. Within the Stealthy family, the additional real CETEM attacks provide an external structural reference, showing how the generated CICIoT23 and Bot-IoT attacks relate to other traffic patterns from the same family. These embeddings complement the quantitative results by showing preservation of both real--synthetic local agreement and broader attack structure.

Figure~\ref{ids_results} shows the downstream IDS performance under real-only training, partial LaDIS augmentation, and complete balancing. Partial augmentation provided the best performance for both classifiers, increasing balanced accuracy and macro F1 while preserving the already high performance of the high-volume families. The largest gains occurred for the low-volume families, whose F1-score increased from 0.73 to 0.84 for TabNet and from 0.82 to 0.88 for XGBoost. In contrast, complete balancing provided no additional benefit and substantially degraded TabNet performance, particularly for the low-volume families. These results indicate that LaDIS-generated observations are effective for reinforcing underrepresented classes without enforcing a uniform training distribution, with consistent behavior across both classifiers.

\section{Discussion}

LaDIS provided a consistent compromise across synthetic data fidelity, structural preservation, real--synthetic distinguishability, and computational efficiency, although no generator was uniformly superior across all metrics and attacks. The results particularly highlight that synthetic data quality cannot be characterized through marginal agreement or manifold coverage alone. High coverage may coexist with distorted marginal distributions, local geometry, or feature dependencies, whereas high-fidelity samples may still provide incomplete coverage. The complementary evaluation adopted in this work therefore provides a more informative assessment of network-traffic synthesis than any individual metric.

The behavior remained attack dependent. For DoS TCP, LaDIS preserved realistic local geometry and dependencies but provided incomplete manifold coverage, whereas Mirai Greip, Recon OS Scan, and MitM ARP Spoofing showed substantially stronger agreement across multiple fidelity criteria. In contrast, TabDDPM frequently achieved very high recall while exhibiting distorted neighborhood scales or marginal distributions, illustrating that broad coverage does not necessarily imply realistic synthesis. Together with the latent dimensionality analysis, these findings motivate selecting representation capacity independently for each attack based on synthetic data quality rather than reconstruction error alone.

Preserving multivariate dependencies is particularly important in network traffic, where realistic flows arise from relationships among multiple descriptors rather than from their marginal distributions independently. The quantitative dependency metrics and CE maps consistently showed that LaDIS reproduced these relationships more accurately for most representative attacks. The hierarchical manifold analysis further supported this behavior by showing that generated observations generally occupied the corresponding real data regions while retaining distinctions between individual attacks. Learned statistical dependencies should, however, be distinguished from deterministic relationships explicitly imposed during reconstruction.

The computational analysis also supports the motivation for performing diffusion in a compact latent representation. Compared with TabDDPM, restricting the iterative reverse process to the latent space substantially reduced sampling time, while AE decoding introduced negligible additional overhead. Training efficiency depended on dataset size, with LaDIS slower than TabDDPM for the largest datasets but faster for the smaller ones, whereas CTGAN and TVAE required considerably longer training times. Although direct CTGAN and TVAE sampling remained slightly faster, LaDIS offers a useful balance between training cost, sampling efficiency, and generative quality for applications requiring repeated generation or model retraining.

The downstream experiments demonstrate that the practical value of synthetic traffic depends not only on its fidelity but also on how it is incorporated into the learning problem. Partial LaDIS augmentation improved balanced accuracy and macro F1 for both TabNet and XGBoost, with the largest gains occurring in the low-volume attack families while high-volume performance remained strong. In contrast, complete balancing provided no additional benefit and substantially degraded TabNet performance. These findings indicate that synthetic augmentation should reinforce underrepresented classes rather than necessarily enforce a uniform training distribution. The agreement between a neural tabular model and a gradient-boosted tree model further suggests that this effect is not specific to a single classifier.

This behavior is relevant for practical IDS development, where the challenge is often not the availability of network traffic itself but the acquisition of sufficiently representative and accurately labeled malicious observations. Privacy and security constraints can limit access to realistic traffic, accurate labeling is labor-intensive, and representative benchmark construction may require attacks to be executed under controlled conditions. Synthetic generation can therefore complement scarce attack data without requiring repeated controlled collection of malicious traffic. In an operational pipeline, LaDIS could be used offline to generate selected underrepresented attacks for classifier training, while validation and testing remain entirely real and the generative model stays outside the online inference path. The present results support selective augmentation driven by downstream utility rather than class equalization as an objective in itself.

Despite these benefits, several limitations should be acknowledged. Detailed generative comparisons were restricted to representative attacks from controlled benchmark datasets, and the downstream experiments do not establish generalization to unseen networks, domains, or attack mechanisms. LaDIS models flow-level observations rather than temporal traffic sequences, and maintaining independent attack-specific generators may introduce scalability and management costs as the number of attacks increases. The present study also does not provide formal privacy guarantees or explicitly quantify memorization risk. Future work should therefore investigate cross-dataset and operational validation, temporal traffic generation, privacy-aware synthesis, and shared generative backbones with attack-specific adaptation.

\section{Conclusions}

This study introduced LaDIS, a latent diffusion framework for attack-specific synthetic IoT traffic generation, and evaluated whether operating in a compact learned representation can provide high-quality, computationally efficient, and practically useful synthetic data. The results support these objectives. Across representative attacks, LaDIS provided a competitive balance of marginal fidelity, multivariate structure, dependency preservation, and real--synthetic indistinguishability, while hierarchical manifold analysis showed that generated samples generally preserved both real data regions and attack-specific organization.

The computational evaluation further demonstrated the benefit of latent space diffusion, substantially reducing sampling cost relative to direct-space TabDDPM while introducing negligible decoding overhead. Finally, downstream experiments with TabNet and XGBoost showed that LaDIS-generated observations can improve IDS performance when used selectively, with partial augmentation increasing balanced accuracy and macro F1 and particularly benefiting low-volume attack families. In general, our findings demonstrate that latent diffusion is a viable strategy for generating realistic IoT attack traffic and that its greatest practical value lies in targeted augmentation of underrepresented attacks rather than indiscriminate class balancing.



\section*{Acknowledgments}
This work was partially supported by CyberFold (ETD202300129) from the European Union (EU NextGeneration, PRTR) through INCIBE; by LATENTIA (PID2022-140786NB-C31), HERMES (PID2023-152331OA-I00), and AI-LINK (AIA2025-163540-C31), funded by MICIU/AEI/10.13039/501100011033 and by FEDER/UE; and by the ELLIS Madrid Node (Autonomous Community of Madrid).

\bibliographystyle{unsrtnat}
\bibliography{thebibliography}

\end{document}